\documentclass{mva_style}
\usepackage{graphicx}
\usepackage{amsfonts}
\usepackage{amsmath}
\usepackage{bm}
\finalcopy 

\begin{document}
\title{An X3D Neural Network Analysis for Runner's Performance Assessment in a Wild Sporting Environment}

\author{
  David Freire-Obreg\'on, Javier Lorenzo-Navarro, Oliverio J. Santana, \\Daniel Hern\'andez-Sosa and Modesto Castrill\'on-Santana\\
  Universidad de Las Palmas de Gran Canaria\\
  {\tt david.freire@ulpgc.es}\\
}

\maketitle

\section*{\centering Abstract}
\textit{
We present a transfer learning analysis on a sporting environment of the expanded 3D (X3D) neural networks. Inspired by action quality assessment methods in the literature, our method uses an action recognition network to estimate athletes' cumulative race time (CRT) during an ultra-distance competition. We evaluate the performance considering the X3D, a family of action recognition networks that expand a small 2D image classification architecture along multiple network axes, including space, time, width, and depth. We demonstrate that the resulting neural network can provide remarkable performance for short input footage, with a mean absolute error of 12 minutes and a half when estimating the CRT for runners who have been active from 8 to 20 hours. Our most significant discovery is that X3D achieves state-of-the-art performance while requiring almost seven times less memory to achieve better precision than previous work. 
}

\section{Introduction}

With the progress of technology, the world of sports undergoes a tremendous transformation as competition teams seek new ways to gain an advantage. Computer vision, which employs artificial intelligence algorithms to analyze camera footage in real-time, is one of the most promising research areas on this topic. In this regard, computer vision has already been utilized in various applications, such as player position estimation, ball trajectory prediction, and technological assistance to referee decisions \cite{Naik22}.

Recently, algorithms for action quality assessment (AQA) have emerged as a result of human action recognition research \cite{Shunli21}. AQA aims to design a system that can automatically and objectively evaluate specific human actions based on input videos. Contrary to the traditional video action recognition problem, AQA evaluates the execution of an action. 

\begin{figure}[bt] 
\begin{minipage}{1\linewidth}
    \centering
    \includegraphics[scale=0.5]{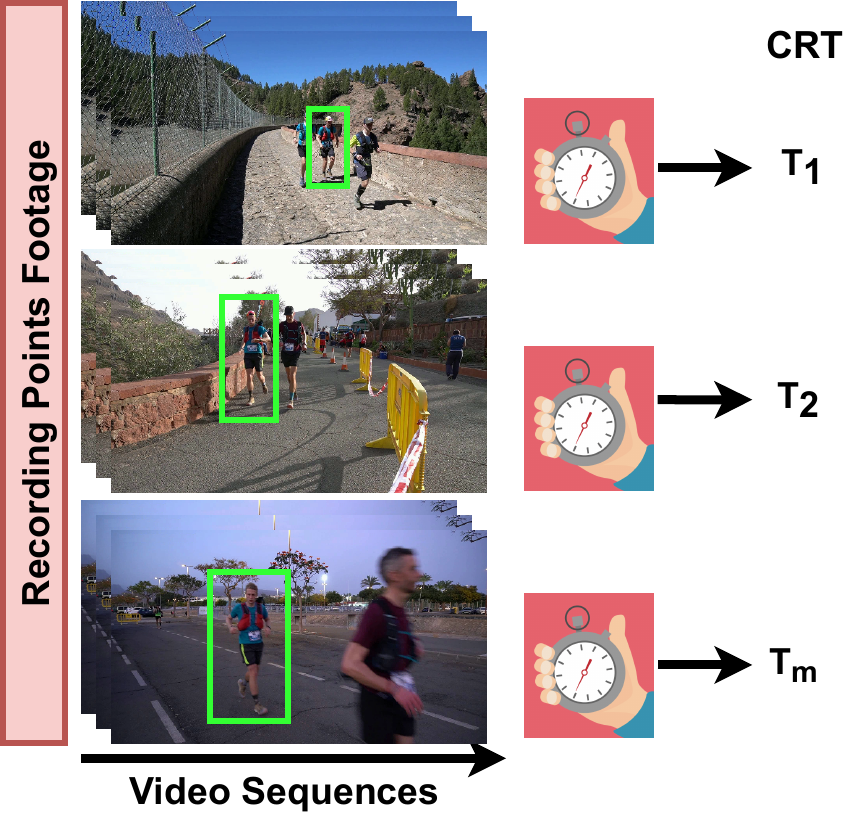}
    \caption{\textbf{Samples of a runner's footage at each recording point.} The runner of interest is surrounded by a green container. We analyze different X3D instances for each footage to extract the runner's embeddings. Then, these embeddings are fed into a model to infer the CRT at a specific recording point.}
    \end{minipage}
     \label{fig:problem}
\end{figure}

Sports have benefited from AQA in many practical scenarios, such as athlete posture correction, coaching systems, and action evaluation. In recent years, the score to be assigned to an athlete's performance by a panel of judges has been estimated, such as diving and gymnastics movements. Consequently, numerous AQA approaches treated this task as a regression problem to learn the direct mapping between videos and action scores \cite{Parmar19, Xumin21}. 

\begin{figure*}[t]  
\begin{minipage}[t]{1\linewidth}
    \centering
    \includegraphics[scale=0.65]{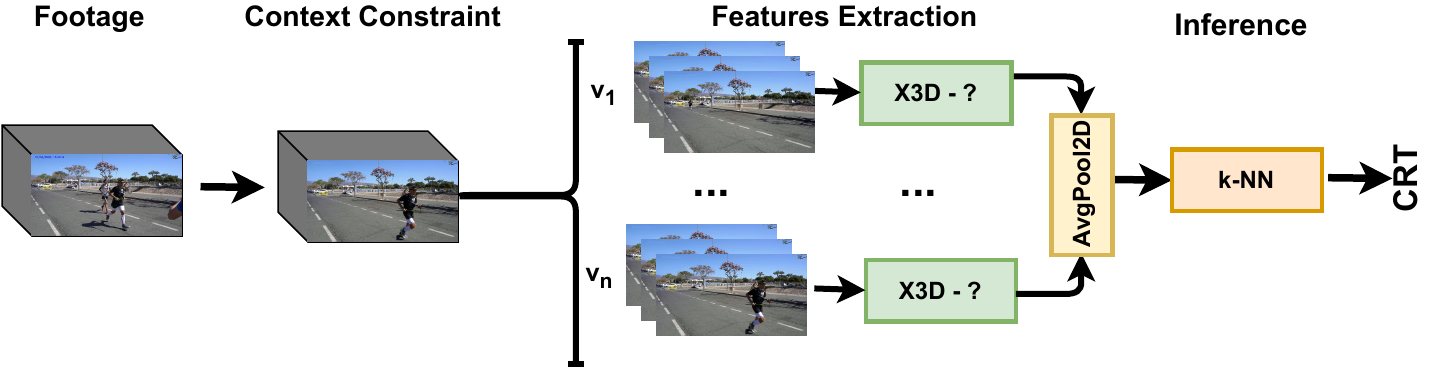}
      \end{minipage}
    \caption{\textbf{The proposed pipeline for regressing runner's CRT.} The designed process consists of two primary components: the footage pre-processing block, and the regression block. In the first scenario, the tracker aids by neutralizing the runner's background activity. The latter entails the division into $n$ small clips by down-sampling. Then the clips are sent into X3D instances for extracting features. An average pooling synthesizes the final features. The resulting tensor is the input to the regressor.
     \label{fig:pipeline}}
\end{figure*}

Lately, ultra-distance competitions have been considered for runners' performance evaluation \cite{freire22iciap}. Contrary to previous sporting AQA works, the task is not to measure the performance of methods between the ground truth and a predicted score series but the CRT. The CRT at a particular recording point $RP_{i}$ can be defined as $T_{i} = T_{1} + \sum_{j=2}^{j=i} (T_{j} - T_{r}) \; where \; r=j-1$, see Figure \ref{fig:problem}. Moreover, the problem framed in ultra-distance races is challenging due to the highly dynamic scenes and the race span, i.e., runner's appearance variance, multiple scenarios, occlusive elements, etc. 

Recent advances in ultra-distance races CRT estimation provide high generality regarding the action recognition networks considered \cite{freire22iciap, freire22icpr}. We seek to bridge these approaches by gradually increasing the network's complexity to resolve this task. Our work explores X3D expanded instances to produce accurate CRT predictions, as shown in Figure \ref{fig:problem}. The used architecture is X3D for expanding from the 2D space into the 3D space-time domain \cite{Feichtenhofer20}. The 2D base architecture is the MobileNet. The considered expansion then progressively increases the computation by expanding only one axis at a time, i.e., frame rate, sampling rate, footage resolution, network depth, number of layers, and number of units. 

We have analyzed the X3D instances in a dataset collected to
evaluate runner re-identification methods in real-world scenarios. The achieved results are
remarkable (up to 12 minutes and a half of MAE), and they have also provided interesting insights. Contrary to other action recognition networks, X3D instances generate shorter embeddings. As a consequence, k-NN instance-based approach turns to be enough to tackle the problem efficiently. Second, a few network expansions are enough to achieve the best performance. Finally, our proposal outperforms other approaches in literature.

\section{Related Work}

AQA is inherently an action recognition problem facing challenges like  automatically and objectively evaluating specific actions people complete through input footage. A common approach to handling action recognition in supervised training has been to limit input data to skeleton-based approaches, e.g., detecting human body joints \cite{Yan18,Duan21}. This encourages the supervised network to infer knowledge from a metric scale that may be globally inconsistent in scenarios where the human body shifts rapidly, such as violence detection \cite{freire22mvap} and sports AQA \cite{Tang20}. In addition, the estimated skeleton data can frequently be noisy in realistic scenes due to occlusions or changing lighting conditions~\cite{Carissimi18}, particularly in ultra-distance races held in uncontrolled environments.

Regarding this issue, appearance-based approaches have been used in the past to tackle AQA. Pioneer research conducted by Parmar et al. already uses C3D neural networks at a clip level for feature computations \cite{Parmar17,Parmar19wacv}. More recently, several works have used the I3D network on clips not to predict a score but a score distribution \cite{Tang20, Zhang21}. I3D ConvNets have also been used to tackle the runner's performance in the past \cite{freire22iciap, freire22icpr}. Contrary to these works, we aim to analyze the athletes performance by progressively expanding an X3D architecture to achieve a lightweight architecture that preserves robustness.

\section{Method}
\label{sec:pipeline}
We develop a modular multi-stage pipeline for runners' CRT estimation in ultra-distance races. Its structure is illustrated in Figure \ref{fig:pipeline}.

\textbf{Context constrain}. According to Freire et al. \cite{freire22icpr}, action recognition networks require clean footage input for CRT inference. Therefore, objects (athletes, race personnel, cars) that are not interesting must be removed from the scene. Specifically, the initial block pre-processes the raw input data to focus on the runner of interest. We have applied ByteTrack \cite{zhang2021bytetrack}, a multi-object tracking network, to track the runner of interest in each footage. Then, a context-constrain pre-processing yielded the scenario considered in our experiments. Given a runner $i$ bounding box area $BB_i(t, RP)$ at a given time $t \in [0,T]$ and in a recording point $RP \in [0,P]$, the new pre-processed footage $F'_i[RP]$ can be formally denoted as follows:
\begin{equation}
\label{eq:contextremoval}
F'_i[RP]=BB_{i}(t, RP) \cup \tau(RP)
\end{equation}

Where $\tau(RP)$ is the average number of frames to generate the clean footage where the runner appears with a still background.

\textbf{Feature extraction and regression}.
The input footage consisting of \textit{n} frames is down-sampled and split into \textit{n} video clips (\textit{{$v_1$, ..., $v_n$}}), each containing $q$ consecutive frames representing an activity snapshot (see Figure \ref{fig:pipeline}). Next, each video clip \textit{vi} is passed through a pre-trained X3D network, resulting in a \mbox{192-dimensional} feature vector. These X3D instances have been pre-trained with the Kinetics dataset \cite{Kay17}, which comprises $400$ action categories. Once all the feature vectors of the \textit{n} video clips are obtained, an average pooling layer is applied to ensure that the information from each clip is given equal consideration. Finally, the extracted features are used to train a k-NN regressor, which is used to infer the CRT.

\textbf{X3D instances}. We have used four X3D expanded instances that are named according to their size; extra small (X3D-XS), small (X3D-S), medium (\mbox{X3D-M}), and large (X3D-L). Each considered expansion is used for sequentially expanding X2D from a tiny spatial network to a spatiotemporal X3D network by performing the following operations on temporal (frame rate and sampling rate), spatial (footage resolution), width (network depth), and depth dimensions (number of layers and number of units) \cite{Feichtenhofer20}. X3D-XS is the output after five expansion steps. The following larger model is X3D-S, defined by one backward contraction step following the seventh expansion step. The contraction step reduces the frame rate and, therefore, temporal resolution while holding the clip duration constant. The eighth and the tenth expansions generate the \mbox{X3D-M} and X3D-L, respectively. As seen in Table~\ref{tab:mae_exp}, X3D-M expands the spatial resolution by increasing the spatial sampling resolution of the input video. In contrast, X3D-L expands not only the spatial resolution but also the depth of the network by increasing the number of layers per residual stage.

\section{Dataset and Experiments}

A key challenge in performing AQA is the lack of publicly available sporting datasets. Our pipeline needs athlete's data to regress CRT properly. While many works in this sporting domain rely on statistical data, manually gathering sufficient multimedia data is expensive. 
We employ a dataset derived from TGC20ReId dataset \cite{penate20prl} provided by the authors, that contains seven-second clips at 25 fps at each recording point for each participant.

The initial dataset includes annotations for nearly 600 participants across six recording points. Given the varying performances of the runners, the gap between the leaders and the last runners increases along the course as the number of active participants decreases. Consequently, a subset of 214 runners is eligible for estimating the CRT, that is, those runners that have covered the last three recording points during the dataset recording time. 

\textbf{Metric}. An athlete $i$ observation $o_i[RP]$ at a recording point $RP \in [0,P]$ consists of a pre-processed footage $F'_i[RP]$ and a CRT $\phi_i[RP]$. In addition, the CRT of the runners has been normalized between [0,1] using Equation \ref{eq:norm}.

\begin{equation}
    \phi'_i[RP] = \frac{\phi_i[RP] - min(\phi_i[0])}{max(\phi_i[RP])}
    \label{eq:norm}
\end{equation}

Our task is to identify an end-to-end regression technique that minimizes the following objective:
\begin{equation}
    min\;L(\phi'_i[RP], \psi_i[RP]) = \frac{1}{N}\sum_{j=0}^{N}|\phi'_i[RP]_j-\psi_i[RP]_j|
    \label{eq:minim}
\end{equation}
where $\psi_i[RP]$ represents for the runner $i$ predicted value at a recording point $RP$ based on seven seconds of movement observation, and $N$ is the batch size.

The following section presents the average Mean Absolute Error (MAE) across 20 repetitions of 10-fold cross-validation.
On average, 410 samples are chosen for training, leaving 46 for testing. 

\subsection{X3D Instances Evaluation}

As Section \ref{sec:pipeline} points out, we have considered several X3D instances, namely X3D-XS, X3D-S, X3D-M, and X3D-L. Additionally, inspired by \cite{freire22icpr}, we have combined these instances averaging them ($192$ embeddings) or concatenating ($192\times\#I$ embeddings, where $\#I$ is the number of combined instances) the last ResNet block output. 

Table \ref{tab:mae_exp} shows the achieved results by each configuration. The table is divided into three blocks, with four entries in the first and three in the rest. The first block is related to the basic X3D instances, the second is the average of different X3D instances embeddings, and the last is the average of different X3D instances concatenation. Average and concatenation experiments combine models sequentially by both, model size -first XS and S, then XS, S and M, and so on- and individual performance. From the temporal dimension perspective, each X3D-XS and X3D-S input clip is composed by four frames and a large sample rate (12 frames), whereas X3D-M and X3D-L increase the number of frames per clip ($13$ and $16$) but reduce the sample rate by a half. 

Table \ref{tab:mae_exp} also highlights the relative importance of the model size. As can be appreciated, smaller models consistently outperform bigger models, i.e., X3D-XS is 18\% better than X3D-L. Averaging model embeddings partially outperform individual approaches, but the rates are inconclusive since there is no correlation between the size and the performance. For instance, the middle combination configuration (XS+S+M) is worse than any other, bigger or smaller, configuration. Freire et al. have recently achieved their best results when concatenating I3D ConvNet embeddings \cite{freire22icpr}. Similarly, in our study, we consistently observed a substantial and consistent reduction in loss as the model size increased, highlighting the effectiveness of concatenating embeddings. Despite these findings, it is worth noting that the X3D-XS model, albeit with a slight margin, still maintains the highest success rate among all tested models.

\begin{table}[!t]
\renewcommand{\arraystretch}{1.3}
\centering
\caption{\textbf{Mean average error (MAE) achieved by each configuration.} The first column displays the model configuration ($+$ stands for average and $\cup$ for concatenation, respectively). The second column shows the number of frames per video clip, the third column shows the sampling rate (SR), and the last column shows the achieved MAE. Lower is better.}
\label{tab:mae_exp}
\begin{tabular}{|l|c|c|c|}
\hline
Instance &  \#Frames & SR & MAE\\
\hline
X3D-XS  & $4$ & $12$ & \bm{$0.010$} \\\hline
X3D-S   & $4$ & $12$ & $0.011$ \\\hline
X3D-M  & $13$ & $6$ & $0.011$ \\\hline
X3D-L  & $16$ & $5$ & $0.012$ \\\hline
\hline
XS+S       & Mixed & Mixed & $0.012$ \\\hline
XS+S+M      & Mixed & Mixed & $0.013$ \\\hline
XS+S+M+L  & Mixed & Mixed & $0.019$ \\\hline
\hline
XS$\cup$S      & Mixed & Mixed & $0.011$ \\\hline
XS$\cup$S$\cup$M     & Mixed & Mixed & $0.011$ \\\hline
XS$\cup$S$\cup$M$\cup$L & Mixed & Mixed & $0.011$ \\
\hline
\end{tabular}
\end{table}

After considering various classifiers such as linear regression, random forest, gradient boosting, SVM, and a multi-layer perceptron, we have found that k-NN outperforms all of them. To ensure the optimal performance, we conducted a grid search to identify the most suitable regressor. It turns out that \mbox{k-NN} reported the best result. Furthermore, due to the small number of dimensions and the moderate number of observations, we have reported rates using a k-NN regression method. Consequently, an instance-based classifier is good enough to select the best embeddings for inference. In terms of minutes, a 0.010 MAE is roughly 12 minutes and a half. Since the fastest runner was recorded after 8 hours of CRT, and the last one after 20 hours of CRT, the achieved MAE is a really positive outcome. 

\begin{table}[!t]
\renewcommand{\arraystretch}{1.3}
\centering
\caption{\textbf{Comparison of different architectures on the dataset used in the present work.} The first column shows the considered pre-trained architectures, whereas the second and the third columns show the number of parameters and the MAE, respectively. Lower is better.}
\label{tab:comp_exp}
\begin{tabular}{|l|c|c|}
\hline
Architecture & \#Params &MAE\\
\hline
C3D \cite{Tran14}      & 34.8M & $0.038$\\\hline
3D ResNets-D30 \cite{Hara18} & 60.5M & $0.036$\\\hline
3D ResNets-D50 \cite{Hara18} &  45.8M    &  $0.033$\\\hline
3D ResNets-D101 \cite{Hara18} &  84.8M   & $0.032$\\\hline
3D ResNets-D200 \cite{Hara18} &  146.4M    &  $0.031$\\\hline
I3D-800SB \cite{freire22icpr} & 25M & $0.019$\\\hline
I3D-2048SB \cite{freire22icpr} & 25M & $0.015$\\\hline\hline
X3D-XS (Ours) &  3.7M   & \bm{$0.010$}\\
\hline
\end{tabular}
\end{table}

To better compare the proposed pipeline with the related work, we have included our best result in Table \ref{tab:comp_exp}. This table summarizes the performance reported in recent literature on the mentioned dataset but also the size of the model. The table includes three major architectures, C3D, 3D ResNets considering different depths, and the I3D ConvNet. Overall, the \mbox{X3D-XS} model outperforms other considered prior architectures on this task. Moreover, Table \ref{tab:comp_exp} shows that the \mbox{X3D-XS} model is more than six and a half times smaller than the model with the second best result. Note that the ranking in this table shows no correlation between the model size and the model performance.

\section{Conclusions}
Combining metric accuracy and lightweight models is a key challenge in AQA. We propose an X3D analysis by progressively expanding the architecture on temporal, spatial, width, and depth dimensions. Then, an instance-based classifier (k-NN) provides good performance on the generated embeddings. We show improved error reduction with each basic X3D instance alone and demonstrate successful results when concatenating instance signals. Our best result was achieved by a model almost seven times smaller and a 34\% better than the best proposal in the literature. Several applications can benefit from our proposal, not only monitoring a runner's performance, but also relieving the race staff from paying exhausting continuous attention to health concerns. In addition, we hope it will assist in deploying robust and general CRT estimation models.

\begin{description}

\item[Acknowledgments:]
  This work is partially funded by the the Spanish Ministry of Science and Innovation under project \mbox{PID2021-122402OB-C22}, and by the ACIISI-Gobierno de Canarias and European FEDER funds under project, \mbox{ProID2021010012}, ULPGC Facilities Net, and Grant \mbox{EIS 2021 04}

\end{description}

\bibliographystyle{IEEEtran}







\end{document}